%% file: main.tex
\documentclass[acmlarge,review]{acmart}

\setcopyright{none}

\acmDOI{}
\acmISBN{}

\newcommand{\myconfshort}{\shortauthors}

\makeatletter

\AtBeginDocument{%
  \fancypagestyle{firstpagestyle}{%
    \fancyhf{}%
  }%
  \fancyhf{}%
  \fancyhead[LO]{\@headfootfont\shorttitle}%
  \fancyhead[LE]{\@headfootfont\footnotesize \myconfshort}%
  \fancyhead[RO]{\@headfootfont\footnotesize \myconfshort}%
  \fancyfoot[C]{\@headfootfont\thepage}
}
\makeatother

\usepackage{array}
\usepackage{enumitem}
\usepackage[title]{appendix}
\usepackage{graphicx}
\usepackage{float}

\begin{document}

\title{Whose Transparency is it Anyway? Understanding Generative AI Transparency Needs and Gaps Across Stakeholders}

\author{Genevieve Smith}
\email{genevieve.smith@stanford.edu}
\affiliation{%
  \institution{Stanford University}
  \city{Palo Alto}
  \country{USA}
}

\author{Sahiba Chopra}
\affiliation{\institution{UC Berkeley}\city{Berkeley}\country{USA}}

\author{Natalia Luka}
\affiliation{\institution{UC Berkeley}\city{Berkeley}\country{USA}}

\author{Min Kyung Lee}
\affiliation{\institution{UT Austin}\city{Austin}\country{USA}}

\renewcommand{\shortauthors}{Smith et al.}

\acmArticleType{Research}

\acmContributions{}

\begin{abstract}
\section*{Abstract}
Transparency has become a central principle in AI governance, yet there is little agreement about what transparency means, for whom, and in what contexts. Drawing on interpretive flexibility within the Social Construction of Technology, this paper examines how different stakeholders understand and experience transparency in generative AI. This study draws from a global survey of 400 end-users, AI builders, and policymakers, complemented by a structured focus group with end-users, grounding the quantitative findings in qualitative insights from the group most underserved by current transparency practice. We present three key findings. First, all three groups share similar understandings of what transparency means and what it should afford, centered on ensuring accuracy and privacy. Second, gaps between the importance of transparency features and their perceived availability are substantially larger for end-users than builders, revealing how power shapes what gets built and who it serves. Only 28\% of end-users report that current transparency practices reflect their needs very well, compared to 53\% of builders and 49\% of policymakers. Third, transparency is situationally constructed, priorities shift with context and task structure, with role-specific frames activating when context demands them and receding when it does not. We argue that meaningful AI transparency requires moving beyond a one-size-fits-all solution, including current emphases on AI disclosure and static documentation, toward pluralistic transparency infrastructures that reflect the needs of diverse users and stakeholders. 
\end{abstract}

\vfill
\enlargethispage{2\baselineskip} 
\vspace*{-1.5pc}                 


\maketitle

\section{Introduction}
\label{sec:introduction}
As generative AI (genAI) becomes increasingly embedded in everyday life, questions about how these systems operate, what data they draw upon, and how their outputs are produced have become matters of urgent public and policy concern. At the center of these conversations is the concept of transparency, broadly understood as the degree to which the design, development, and deployment of AI systems are made visible and understandable to stakeholders \cite{larsson2020transparency}. Transparency has emerged as the most widely cited principle in AI ethics \cite{jobin2019global}, yet a critical gap persists between the principle of transparency and its practice. 

This paper addresses that gap. We argue that transparency is not a single, stable concept but rather an interpretively flexible one that is valued and experienced differently depending on who is doing the asking and why. An end-user trying to understand whether to trust a chatbot and its output, an engineer using foundation models to create new products, and a policymaker seeking to assess societal impact and regulatory compliance occupy fundamentally different epistemic and practical positions. Each brings distinct needs for what counts as meaningful transparency. Current approaches in practice rarely account for this diversity, and as a result, risk serving some stakeholders well while leaving others significantly underserved.

We present findings from a pre-registered survey
\footnote{Pre-registered on As-Predicted:https://aspredicted.org/m5fv8j.pdf}
of 400 participants across three stakeholder groups: end-users of genAI systems, policymakers and policy experts engaged in AI governance, and AI builders (including both product team members and technical developers). Survey findings are complemented by a structured focus group with end-users that deepens and contextualizes insights. Our research investigated how people understand and value genAI transparency, as well as the gaps within and across groups. We draw on interpretive flexibility from the Social Construction of Technology (SCOT) as a sensitizing concept to orient our inquiry into how different stakeholder groups understand and experience transparency in genAI.

Our findings reveal three interconnected patterns. First, all three stakeholder groups share a similar normative foundation for what transparency means and what it should afford, centered on ensuring accuracy and privacy. Second, and despite this convergence, existing transparency features and resources serve builders significantly more effectively than end-users. Only 28\% of end-users report that current transparency practices reflect their needs very well, compared to 53\% of builders and 49\% of policymakers. This pattern is consistent with a development process in which builder communities have substantially more influence over what transparency features and artifacts get built and in what form \cite{suresh2021beyond}. Third, transparency priorities are situationally constructed, shifting with context and task structure, with professional frames activating when context demands them. Together these findings suggest that meaningful AI transparency cannot be achieved through a single standard applied uniformly. Rather, meaningful transparency in genAI requires pluralistic transparency infrastructure accommodating the diverse needs of different stakeholder groups and actively centering those most underserved. 

This paper provides the first systematic cross-stakeholder comparison of transparency understandings and gaps in current implementations of genAI. We further offer concrete recommendations for how genAI transparency should be approached. Findings carry implications beyond academia. Trust in AI systems is declining among the public \cite{gillespie2025trust}. Our findings suggest a reason for this: what people most want when using genAI is to be able to trust the system and what it produces, and they understand transparency as a primary means of establishing and calibrating that trust. Yet current transparency tools are failing to provide transparency mechanisms adequately, particularly for end-users. Addressing transparency gaps meaningfully for different stakeholders therefore is not only a matter of equity and governance, but a practical imperative for those developing these systems.

\section{Background}
\label{sec:related-work}

\subsection{Transparency across the AI lifecycle}
At a high level, transparency involves making the inner workings of AI systems understandable to stakeholders and is considered foundational to building accountability and trust \cite{schmidt2020transparency}. As genAI technologies grow more prevalent and powerful, transparency has become an area of critical importance in AI governance, focusing on openness in the design, development, and deployment of AI systems, ensuring that processes and mechanisms are visible and understandable to those who interact with them. Transparency in AI also includes broader socio-technical dimensions, including the ways AI technologies interact with and impact society \cite{larsson2020transparency}. 

Transparency in AI encompasses multiple interrelated yet distinct facets spanning the AI lifecycle, from upstream, model-level, and downstream domains \cite{felzmann2020towards}.  Upstream transparency includes openness regarding the inputs, assets, and decisions made prior to model training. Transparency in this domain can include disclosing the provenance and composition of training data (such as through tools including datasheets for datasets \cite{gebru2021datasheets}, compute resources, and labor practices. 

At the model level, transparency includes various concepts. Openness at this level includes making model parameters and weights publicly downloadable alongside system architecture to support external auditability and local deployment. Interpretability and explainability help explain the technology itself. AI models, particularly deep learning models including genAI, are inherently complex and operate as “black boxes” making it difficult to understand their decision making processes \cite{felzmann2020towards}. Interpretability research explores the internal representations and decision logic driving model behavior \cite{saeed2023explainable}. Explainability helps bridge the gap when models operate as “black boxes” by providing human-understandable justifications or high-level approximations for why a specific decision was made \cite{mittelstadt2019explaining, saeed2023explainable}. Explainability tools can include model cards \cite{mitchell2019model} and system cards. Post-modeling methods can include textual justification, confidence scores, citations, and benchmarking \cite{minh2022explainable}.

Other forms of downstream transparency informs how AI systems interact with end-users and broader society. Visibility ensures that stakeholders are informed about the presence and role of AI in the first place, such as through disclosures or watermarks on AI-generated content. Accessibility ensures that information is usable and actionable for all relevant stakeholders, such as through ensuring risk evaluations and terms of service are clear across technical levels \cite{felzmann2020towards}. Finally, downstream transparency includes information about societal impacts. 
There remains a lack of transparency from companies developing foundation models, particularly in the United States \cite{burrell2016how, lemley2023how}. The Stanford Foundation Model Openness Index, which has scored foundation model developers since 2023 across 100 transparency indicators, found that companies have not improved over time, with a 17-point average decrease documented from 2024 to 2025 (Stanford Center for Research on Foundation Models, 2025). Training data remains particularly opaque, with little disclosure on copyright and licensing, and many companies still do not share basic information such as model size and architecture. Companies may choose to withhold details about models for various concerns such as competitive advantage, intellectual property, and exposure to other legal or reputational risks such as copyright infringement. Developers may also claim that limited transparency is valuable for safety and security, or privacy concerns, especially in multi-agent systems \cite{felzmann2020towards}.

\subsection{Divergent stakeholder needs and limitations of current approaches}
\label{sec:Divergent}
AI transparency is not a static or globally uniform concept. Rather, different stakeholders can require different types of transparency linked to different purposes (Weller, 2019). For example, developers may seek transparency on internal model workings to enable them to understand, debug, and enhance AI models. Non-technical end-users may be interested in whether they can trust an AI-generated output enough to use it and policymakers seek mechanisms for recourse and accountability. Furthermore, the literature reveals a significant gap between design of transparency artifacts, which are largely developed by and for technical audiences, and utility for diverse audiences \cite{suresh2021beyond}. This is linked to researchers’ developing explainable AI toolings using their own intuitions, versus considering other needs, in what makes a “good” explanation \cite{miller2019explanation, suresh2021beyond}. Suresh et al (2021) further note this gap reflects power dynamics whereby those with formal technical knowledge designing transparency and interpretability systems are often in positions of power. Without involving other stakeholders with different interests and needs, such systems will continue to serve those in power. Even where transparency measures are successfully implemented, their capacity to support accountability depends on the presence of an audience that is able to engage critically with what is disclosed \cite{kemper2019transparent}.

Human-centered explainable AI (HCXAI) literature has historically focused on single-task predictive models deployed in high-stakes domains such as healthcare \cite{cai2019hello} and criminal justice \cite{ghayatrie2026explainable}. Yet even in these domains, existing transparency artifacts fail to satisfy the operational and legal needs of key stakeholders across the technology lifecycle \cite{ghayatrie2026explainable}. These findings mirror broader critiques which emphasize that mainstream explainable AI (XAI) approaches tend to treat the user as an idealized technical peer and often don’t meet their needs \cite{liao2021humancentered, liao2021question}. Exposing non-technical users to dense information about the model can result in information overload \cite{ehsan2021expanding}. Even amongst technical users, existing tools can cause cognitive fatigue and result in misinterpretation. An empirical study evaluating how data scientists interact with popular interpretability tools (e.g., LIME, SHAP), found that visualization outputs often created a false sense of security with over-trust and misuse of interpretability tools \cite{kaur2020interpreting}. For regulators and policymakers, explainability tools including static documentation artifacts do not necessarily provide information required for legal enforcement and public accountability \cite{raji2020closing}. This reflects a more fundamental limitation of static disclosure that captures a momentary snapshot of systems that continue to evolve, undermining their capacity for ongoing accountability \cite{ananny2018seeing}. A systematic audit of over 32,000 model cards further found that information provided in model cards is often lacking and uneven \cite{liang2024systematic}. In addition, current explainability mechanisms create barriers for neurodivergent populations \cite{bennett2021its, nwokoye2024survey}.

While a growing literature acknowledges that different stakeholders have different needs regarding transparency, systematic empirical studies comparing these needs in the context of genAI are lacking. Empirical analyses examining how existing transparency approaches across the AI lifecycle meet the distinct needs of developers, end-users, and policymakers are limited. Moreover, existing research has largely been conducted within North American and European contexts. This study seeks to address these gaps. 

\subsection{Theoretical orientation: Interpretive flexibility and power}
\label{sec:theory}
This study draws primarily on the concept of interpretive flexibility from the SCOT literature. We use interpretive flexibility, which holds that technologies do not have fixed, intrinsic meanings or functions, as a sensitizing concept to orient our inquiry \cite{pinch1984social}. In this, the meaning and workings of technologies are socially constructed through the interactions of relevant social groups or actors who engage with a technology and attribute meaning to it, whereby the same artifact or concept can mean different things to different social groups \cite{bijker2001technology}. Applied to AI governance, what counts as a “transparent” AI system can differ fundamentally, as developers, end-users, and policymakers operate from distinct assumptions, values, and objectives reflecting distinct technological frames that shape how each group thinks about and interacts with a technology. 

Interpretive flexibility does not persist indefinitely. Through processes of stabilization and closure, one interpretation tends to gain dominance and the range of meanings narrows. However, whether closure – which describes the endpoint at which interpretive flexibility effectively vanishes and one construction of the artifact prevails as technologies are deployed and used \cite{bijker2001technology} – is inevitable or even possible in the context of genAI remains an open empirical question. Unlike a bicycle or a lamp, which have relatively bounded use contexts around which a dominant interpretation can crystallize, genAI is simultaneously a medical, educational, creative, and professional tool across an effectively unbounded range of users and contexts. This may create conditions under which any convergence around transparency, both as a concept and as a set of artifacts and practices, may be perpetually deferred. 

Whereas early SCOT has been criticized for insufficient attention to power by treating all social groups as analytically equal thereby obscuring the influence of some groups in stabilization processes \cite{winner1993upon}, we pay particular attention to power. In practice, developers and tech firms – as well as policymakers in some contexts – exert disproportionate influence over which transparency mechanisms stabilize into industry defaults. We direct analytical attention to whose technological frames are currently prevailing, operationalizing this cross-sectionally by mapping gaps between perceived importance and availability of transparency features across stakeholder groups.  

\section{Methodology}
\label{sec:methodology}
\subsection{Study design}
This study draws from a survey of participants across three stakeholder groups to examine how different stakeholder groups understand, prioritize, and experience transparency in genAI systems. Survey findings are complemented by a structured focus group with end-users, the group most underserved by current transparency practice. All study procedures were approved by the Institutional Review Board at [Anonymized University] (Protocol \#[Anonymized]). Informed consent was obtained from all participants prior to data collection.

\subsubsection{Survey}
Survey participants were recruited through the Prolific online research platform, the “friend-of-a-friend” method in which team members reached out to individuals within their networks who met the study's inclusion criteria, and snowball sampling. The recruitment strategy was designed to capture international participation, including through inclusion goals on Prolific for different geographic regions. End-users and builders were primarily recruited through Prolific, while recruitment for policymakers primarily utilized the “friend-of-a-friend” method and snowball sampling (details on inclusion criteria and recruitment in Appendix 1a). Prior to full deployment, the research team piloted the survey with a subset of participants. Survey data was collected from October 2025 through April 2026. Completing the survey took approximately 20 minutes with respondents paid US\$6.25. 
The survey sample is 400 participants distributed across three stakeholder groups: 250 end-users of genAI, 94 AI builders (technical AI developers and product team members integrating genAI into products), and 56 policymakers (individuals working in or advising on AI regulation and policy). This distribution reflects the relative availability of each population for research recruitment while ensuring sufficient representation across groups for meaningful comparison. Respondents were fairly evenly split by male (53\%) and female (46\%), with 1\% of respondents identifying as non-binary, and predominantly identified as White (42\%) or Black/African American (39\%), with smaller shares identifying as Latino/Hispanic, East Asian, South Asian, Indigenous, or Pacific Islander (see Appendix 1b for full demographic breakdown). 

All participants completed an informed consent process before beginning the survey. The survey instrument was organized around three domains: (1) what people want from AI and transparency, including how participants define and conceptualize transparency; (2) transparency importance and gaps, measuring the degree to which specific transparency features are considered important and the degree to which they are perceived as currently available; and (3) context variation, using scenario vignettes to assess whether transparency expectations remain stable or shift across different use contexts and perceived risk levels.

\subsubsection{Focus group with structured activities}
At the close of the survey, participants were invited to indicate interest in taking part in a participatory design process to better understand their needs and build new transparency artifacts. We focused on the stakeholder group that was most underserved, as demonstrated in our survey findings: end-users. From among those who are end-users, self-selected, and provided a valid email, we randomly selected 20 participants maintaining at least 45\% recruitment from both male and female participants, maximum 50\% recruitment from the US, and representation across global regions.

Selected participants were contacted via email to confirm availability and willingness to participate, as well as complete an informed consent process. Participants were paid US\$40 for participation in each session. Participatory design sessions were conducted via Zoom across three two-hour sessions in February and March 2026. This paper draws on the first session, which was conducted as a focus group with structured activities and elicited deeper understandings of transparency needs and gaps. In this initial session, participants were fairly split across male (50\%) and female (45\%), with one non-binary participant. Geographically, 33\% of participants were from outside the U.S. Non-U.S. participants were from Mexico, India, South Africa, Canada, and the United Kingdom.
Participants participated in two facilitated activities. In the first, participants brainstormed what they want from AI transparency, then clustered responses into categories in small groups. In the second, participants read three vignettes in which generative AI produced misleading results, then voted on which transparency features they most wanted and perceived as available. Group discussion debriefed activities (see Appendix 2 for activity descriptions and illustrative outputs).

\subsection{Analytical approach}

\subsubsection{Survey}
Analysis was conducted across the three survey domains. We assessed overall trends across the full sample and then compared across the stakeholder groups. Secondary comparisons were conducted across geographic context, using World Bank income classification through a three-group split (high-income n=241, upper-middle n=147, low- and middle-income n=12). Additional secondary analyses were done based on technical level based on self-reported coding and AI expertise (low n=274; high n=126). 

In the first domain, we explored responses to several questions that examined what people want from AI and transparency. Specifically, what matters most when using genAI tools and what people want transparency to achieve. These questions were assessed via a forced-choice list. In the second domain, we explored the importance of different transparency features and gaps for whether those features that were identified as important were also perceived as available. We assessed 13 transparency features identified through a review of the AI transparency and explainability literature and supplemented through research team discussion to ensure coverage of dimensions relevant to diverse stakeholder groups. Features were categorized as “user-facing” or “governance”. We distinguish user-facing transparency that users may encounter during system use (6 items: AI disclosure labels, plain-language explanations, confidence indicators, limitations on what the system should be used for or not, source citations, system prompts that guide responses), from governance-facing transparency that concerns the structural rules, protections, and oversight mechanisms of systems (7 items: training data information; public documentation on the system, data use, limitations, and risks
\footnote{Documentation resources (e.g., model cards, system cards) are designed to consolidate many governance transparency features in one place (e.g., training data, performance evaluation, limitations, societal impact). These features can also exist outside of formal documentation through other channels, such as independent audit reports or corporate responsibility disclosures, though documentation tools represent a primary mechanism for governance transparency in the field. We therefore treat documentation as a distinct transparency feature alongside the specific governance features it is designed to consolidate, allowing us to examine whether awareness and availability of the consolidated resource differs from awareness of its component parts.}; 
public reporting on social and environmental impacts; independent audits; benchmarks; privacy policies; recourse mechanisms). These features were rated on a 5-point Likert scale for importance, and a yes/no/unsure measure of perceived availability in the latest genAI tool they used or one they regularly use. To assess gaps between importance and availability, importance ratings were rescaled to a 0-100 scale and availability expressed as the percentage answering “yes” among those giving a “yes” or “no” answer, placing both measures on a common metric. The gap for each feature is the difference between these two quantities, computed separately within each stakeholder role and within each secondary subgroup. This approach adapts the logic of importance-performance analysis \cite{martilla1977importance}. 

Finally, we explored how transparency needs vary in different contexts and among stakeholders. This was examined through four scenario vignettes varying in stakes and role specificity. A low-stakes information task not specific to any stakeholder role (i.e. “neutral” scenario) was presented first, followed by a higher-stakes role-specific scenario for each stakeholder group. Participants ranked transparency features by usefulness for each vignette, with priority shifts compared between each scenario. 

\subsubsection{Focus group analysis}
Data from the focus group with structured activities consisted of two sources: workshop artifacts (participants' sticky note brainstorms, clustered categories, value associations, dot voting results), as well as session transcripts from the group discussions. Both sources were compiled into a Word document and coded qualitatively using NVivo. Top-level codes were organized around the survey domains (transparency needs and meanings, importance and gaps, context variation). Within each domain, an initial set of subcodes was derived deductively from the survey findings (e.g., explainability, sources and citations, accuracy, privacy, and accountability within the transparency needs domain). Additional subcodes emerged inductively. One team member conducted an initial round of coding, after which the full codebook was reviewed and refined by the research team before a final coding round.

\section{Findings}
\label{sec:findings}

\subsection{What people want from AI and transparency}
\label{sec:wants}
We begin by examining what respondents most value when using genAI, as well as what they want transparency to achieve. When asked about the two factors that matter most when using AI tools, trusting it’s true or accurate and protecting privacy and personal data emerged as the top two factors across all three groups (see Figure 1 in the Appendix). Some significant differences emerged. End users were significantly more likely to prioritize trusting that outputs are true or accurate than both builders and policymakers (p<.01), and more likely to want assurance it’s fair and unbiased than policymakers (p<.05). Builders were significantly more likely than end-users to prioritize understanding how it works (p<.001). Policymakers were significantly more likely than both end-users and builders to prioritize knowing an output is AI-generated (p<.01). 

When asked what they want transparency to achieve, protecting privacy and sensitive information was one of the top two goals for each group. Ensuring AI systems are safe, secure, and reliable was also in the top two for end-users and policymakers, while ensuring fairness came in second for builders. The latter is notable given that builders did not prioritize fairness as a top value when using AI tools (see Figure 1 in the Appendix), suggesting that fairness may be a higher priority when developing systems for others, reflecting a responsibility to ensure tools work equally well and are reliable for different groups. Taken together, people most want to know genAI systems are accurate and trustworthy, with transparency being a way to help them get there, including through ensuring interactions protect privacy and support reliability. 

\begin{figure}[htbp]
    \centering
    \includegraphics[width=0.7\textwidth]{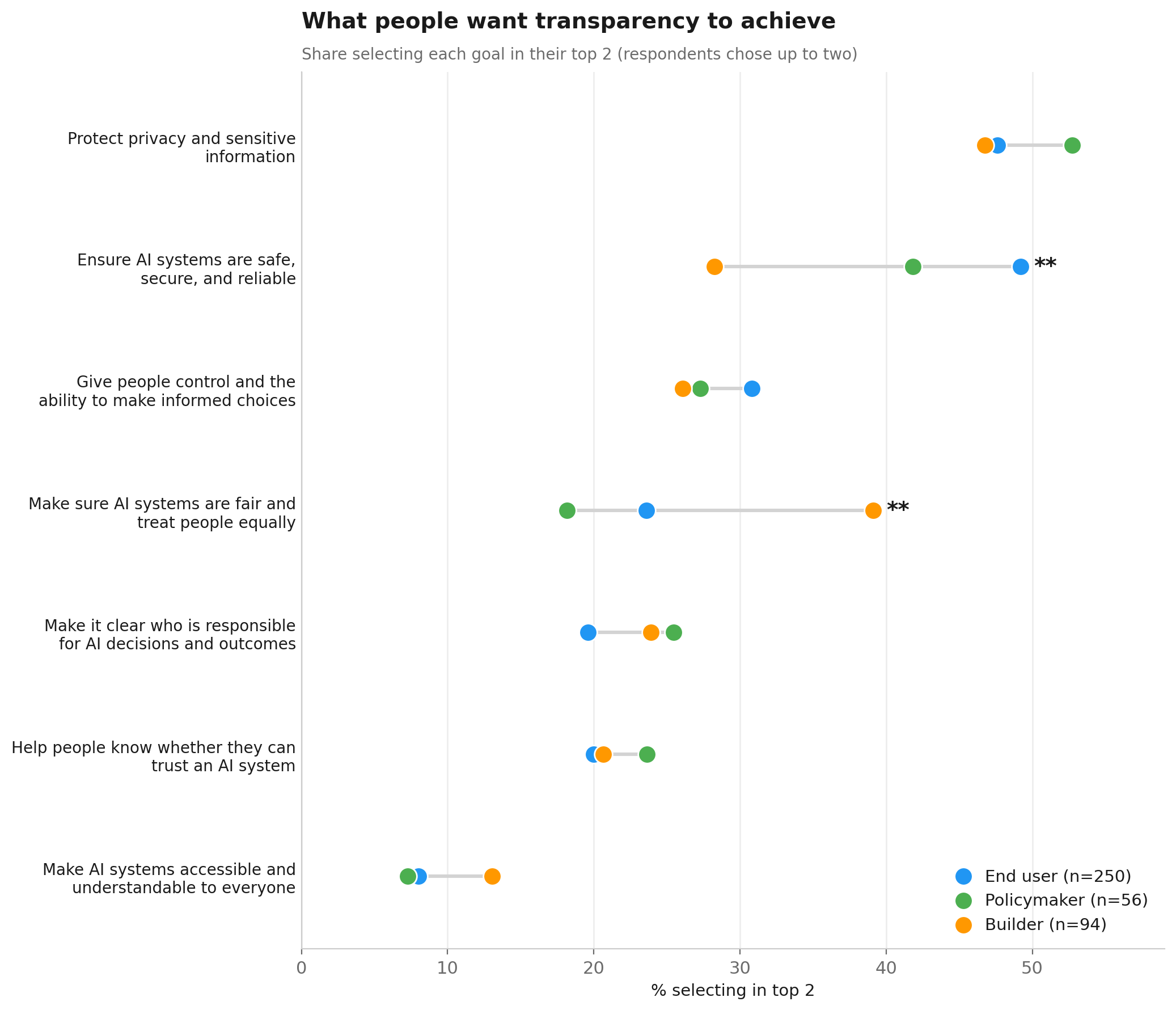}
    \caption{What goals transparency should achieve, by stakeholder role.}
    \label{fig:fig1}
    \vspace{2pt}
    {\scriptsize\raggedright \textit{Note.} Respondents chose up to two options in response to: "Which of the following do you think are the most important goals of transparency?" Stars mark an omnibus Pearson $\chi^2$ test (goal selected vs. not x three roles, df = 2): * p < .05, ** p < .01, *** p < .001. Significant omnibus differences were followed up with pairwise Fisher’s exact tests reported here (fairness: B vs. EU p = .006, B vs. PM s p = .010; safety and reliability: EU vs. B p = .001). .\par}
\end{figure}

\subsection{Transparency importance and gap analysis}
\label{sec:importance}

\subsubsection{Transparency importance}
Importance ratings across 13 transparency features largely corroborate the priorities established while making them more precise and operationalizable. Across all groups, every feature was rated as important, with mean scores ranging from 3.97 to 4.75 on a 1 to 5 scale across stakeholder groups, with 1 being not at all important and 5 being extremely important (see Figure 2 in the Appendix). Citations and sources and privacy information were rated highly across all three groups, consistent with ensuring accuracy and privacy as shared top needs. Where groups diverged, patterns reflected their broader profiles. End-users rated citations and sources significantly higher than builders and policymakers (p<.001). Policymakers rated governance features significantly higher than other stakeholders, including training data information (p<.001), societal and ethical impact (p<.01), and independent audits (p<.05). Builders rated usage limits significantly higher than end-users among user-facing features (p<.05). 

Findings from the focus group with end-users reflected these patterns. When asked what they want from transparency, end-users returned to themes of ensuring accuracy, calibrating trust, and understanding privacy protections (or lack thereof). Transparency features consistently identified for ensuring accuracy and calibrating trust included sources, citations, and confidence scores. One participant painted the connection clearly, “We want to know the sources so that we can make sure that the information is accurate”. Regarding privacy, participants consistently discussed desiring transparency around how user data – particularly sensitive prompts – was accessed, retained, and used. One participant reflected on their group’s discussion: “Privacy seemed to be a big concern. Obviously you don't want AI to retain your problems if it's something personal.” Discussions on privacy reflected a desire to trust not just AI outputs but the organizations providing them. One participant shared how their group wanted to understand, “Where [my] data is going, where it’s being stored, and what it is being used for.” They continued, “Who is going to hold them responsible? Who are they accountable to for this information that they are generating and [the data they] are saving?” Participants recognized the opacity of tech companies developing AI models and several voiced concerns that present-day regulations were too weak to hold companies producing AI systems accountable. The structured focus group reflected how end-users value transparency instrumentally, as a means through which accuracy and trust become verifiable – or not – at both the output level and broader system level. 

\subsubsection{Transparency gaps}

\paragraph*{Major unmet needs, particularly for end-users.} 
Across the full sample, transparency features were rated as highly important but perceived as widely unavailable. Gap scores – the difference between mean importance (rescaled 0-100) and perceived availability (\% reporting the feature as available) – ranged from 17 to 46 across all 13 features. No feature showed a gap below 10. This indicates that perceived importance exceeded perceived availability for every transparency feature examined overall, suggesting transparency is not currently well served across the sample as a whole.
\footnote{Item non-response was concentrated in the governance block: 121 respondents could not say whether independent audits were available, 112 for training data information, 97 for documentation, and 96 for societal and ethical impact, compared with 10 to 35 for user-facing features.}

Gaps differed substantially across stakeholder groups. End users experienced the largest gaps overall (mean gap=35 points) at nearly twice the gap of builders (mean gap=17 points; p<.001). With policymakers in between (mean gap=22 points) (see Figure 3 in the Appendix). This pattern is corroborated by participants’ subjective assessments of how well current transparency practices reflect their needs, which differed significantly across stakeholder groups ($x^2(4)=24.5$, p<.0001). Only 28\% of end-users report current practices reflect their needs very well compared to 53\% of builders and 49\% of policymakers. The correlation between subjective needs alignment and mean gap score is strong and significant ($\rho=-0.50, p<.0001$).

Secondary analyses examined whether gap patterns varied by gender and geographic income context. Mean gaps across the 13 features were 27.5 for men and 28.2 for women, and differences are not significant. As a post-hoc exploration, gaps appear descriptively larger among respondents from low- and middle- income countries (n=12), with an overall mean gap of 34 compared to 30 for high-income countries (n=241), though this difference is not statistically significant (p=0.70). Given the small sample, this observation is speculative. Future research with larger and more geographically balanced samples is needed. 

Regarding why gaps may be particularly large for end-users, documentation provides an instructive case. Despite citing documentation (“public documentation on how the system was built, what data it used, its limitations, and known risks”) as important, end-users consistently report it as unavailable (gap score of 38). A survey question that asked about awareness and usefulness of popular documentation tools (incorporated given the prominence of documentation tools in AI transparency discussions and policy) reveals why. Only 29\% of end-users are aware of or have used system cards compared to 73\% of builders and 68\% of policymakers, and only 32\% are aware of model cards (vs. 76\% of builders, 70\% of policymakers). Among those who are aware, end-users also rate these tools as less useful (e.g., system cards rated 69 out of 100 for usefulness among aware end-users, compared to 80 for builders). Only a small fraction of end-users are both aware of and find these tools useful, compared to a majority of builders and policymakers. This illustrates how governance-related transparency features, including primary documentation tools that the field has aggregated around, are less visible to end-users generally and less legible when they are encountered; thus contributing to larger gap scores. 

\paragraph*{Universal unmet needs across groups.}
Several features show large gaps across all three stakeholder groups, suggesting they represent universal unmet needs. Confidence tagging, training data information, and system prompts appear among the largest gaps for end-users, builders, and policymakers alike. Confidence tagging is particularly striking – with gaps of 55 for end-users, 29 for builders, and 30 for policymakers – indicating that the ability to know how certain an AI system is about its outputs is both broadly desired and unavailable (see Figure 2). End users report the lowest availability across 12 of the 13 features.

\begin{figure}[htbp]
    \centering
    \includegraphics[width=0.7\textwidth]{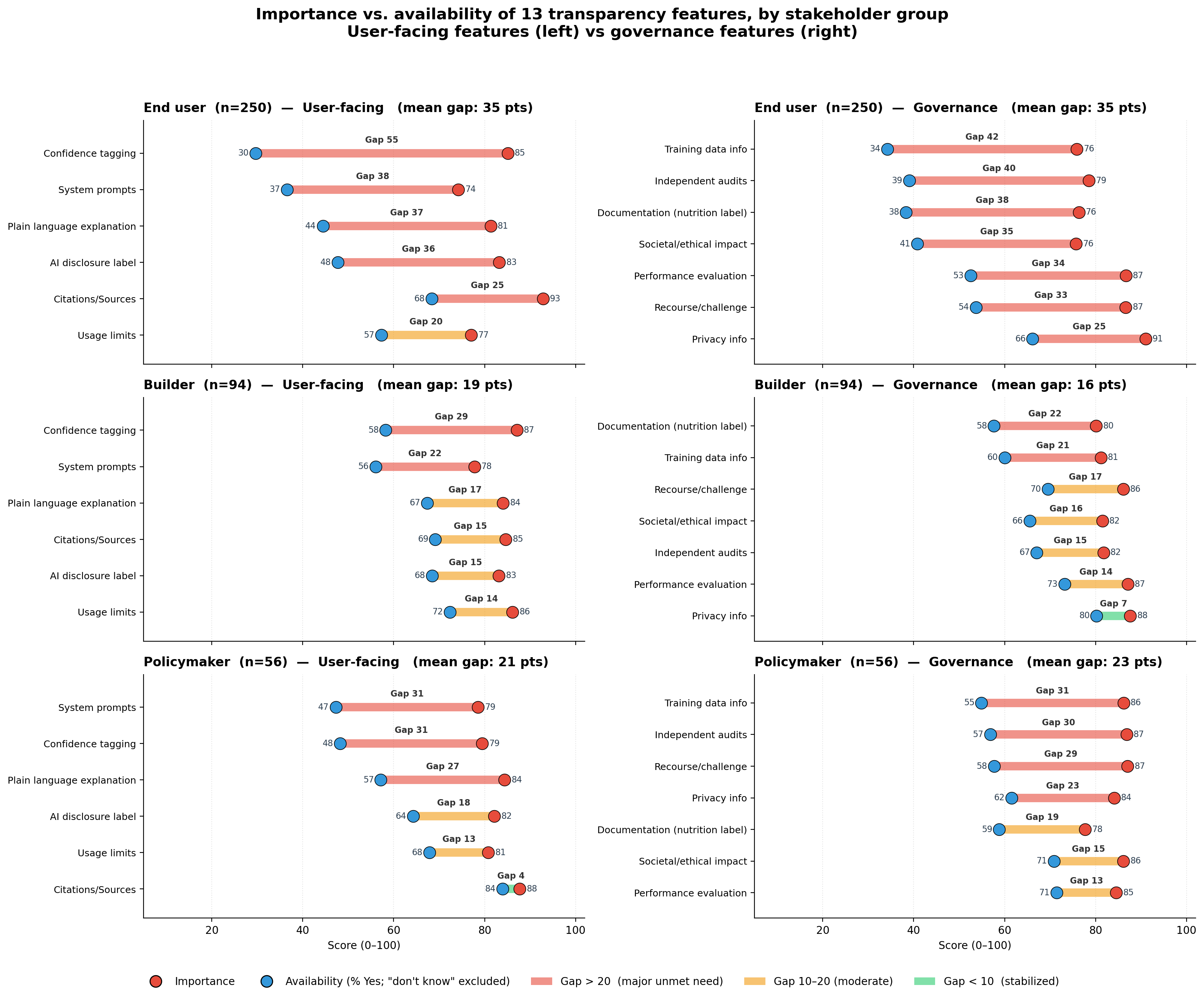}
    \caption{Importance versus availability of transparency features, by stakeholder group}
    \label{fig:fig5}
    \vspace{2pt}
    {\scriptsize\raggedright \textit{Note.} Note: Figure 2 plots rated importance (mean rating rescaled to 0–100) against reported availability (\% answering yes among those answering yes or no) for each transparency feature, by stakeholder role. Bar colors indicate gap magnitude: dark red = gap > 20, light red/orange = gap 10–20, green = gap < 10. Importance is uniformly high across roles: end-users M=81.9 (SE=0.7), builders M=83.7 (SE=1.1), policymakers M=83.2 (SE=2.2). Availability is not: 46.9 (SE=1.8), 65.1 (SE=3.1), 60.2 (SE=4.2). End-users report the largest gaps: M=35.6 (SE=1.9) versus builders 19.4 (SE=3.2) and policymakers 23.0 (SE=4.4). "I don't know" responses are excluded from the availability denominator given various potential meanings eg., never looked for the feature, did not understand it, could not recall it. \par}
\end{figure}

\paragraph*{Some stabilization is occurring.}
While no feature has stabilized across the full sample (whereby importance and availability align), examining the lowest gaps by group reveals that some partial stabilization is beginning to occur for some groups. For builders, privacy information is the only feature approaching stabilization (gap=7), with eight other transparency features in a moderate range (see Figure 2. For policymakers, citations and sources show the smallest gap of any feature across any group (gap=4), suggesting this feature is fairly stabilized for the regulatory community, with five other features in a moderate range. For end-users, no feature approaches stabilization, with usage limits as the smallest gap still at 20 points. Where some stabilization is occurring it is local to particular groups and differs across them (i.e. privacy info for builders, citations and sources for policymakers) rather than representing convergence across stakeholders.

Finally, gap size alone should not be interpreted as a straightforward guide to prioritization. Features with smaller gaps are not necessarily well served; rather they may be more available relative to others while still falling far short for stakeholder needs. Privacy information, for instance, shows a gap of 25 for end-users, one of the smallest gaps though also among the most important features desired. 

\subsection{Context variation: Task structure and stakes shape transparency priorities}

Survey participants saw the same neutral low-stakes scenario and then a higher-stakes scenario specific to their role: a health query for end-users, a policy-briefing query for policymakers, and a model-procurement decision for builders (Appendix 3). In each scenario, participants chose the three transparency features they would find most useful (see Figure 4 in the Appendix). In the low-stakes vignette the top three most-selected features for each stakeholder group include sources, explanation, and confidence. Within those, some stakeholder differences emerge with end-users selecting sources more often (72\%) than policymakers (55\%) or builders (48\%; x², p<.001). 

Comparing low- and high-stakes scenarios reveals that transparency priorities are both partially stable and partially situationally constructed. Sources, explanation, and confidence tagging remain valued across all four scenarios, the most stable features in the survey data. Findings from the focus group with end-users reflect these findings: The most prominent features identified and discussed across both the low-stakes scenario and high-stakes scenario were sources, explanations in everyday terms around how the output was generated, and confidence scores. Focus group participants also identified limits as particularly important in the high-stake case, which was a similar priority for end-user survey participants. One focus group participant explained that in these high stakes scenarios, “the system should make clear what it should and what it should not be used for.” Limits help to avoid overreliance when outputs carry real consequences.

In the survey data, usage limits show the most dramatic sensitivity to context. Among end-users, selection of limits rose from 12\% in the neutral low-stake scenario to 44\% in the health scenario (+32 points, p<.001), while it also rose significantly in the builder scenario from 15\% to 38\% (+23 points, p<.001). This reflects how limits serve different purposes across these contexts. In the health scenario, understanding limits functions as a means of calibrating personal trust, as illustrated by the focus group quote, helping users know when to rely on AI and when to defer to professional expertise. In the builder scenario, limits may function as a means of helping product teams define appropriate scope of model use before deployment.

The builder scenario produced the most distinctive profile. It was the only scenario that asked participants to choose between AI models rather than evaluate an output, which is reflected in their choices shifting toward governance features. The share placing at least one governance feature in their top three rose from 56\% to 77\%, and the mean number selected rose from 0.85 to 1.19 (Wilcoxon p<.001). No single governance feature shifted significantly on its own. Builders' lower selection of sources (43\%) did not reflect a shift, as they had also selected sources less than other groups in the neutral scenario (48\%). 

The policy scenario, by contrast, remains similar to the neutral low-stakes scenario despite its high-stakes content. While the scenario was designed to be relevant for policymakers (involving research into AI transparency standards), it still placed policymakers in an output evaluation role of assessing a chatbot response versus making a regulatory decision. Therefore, it may not have fully activated their professional governance frame. Together these patterns suggest that the task structure is an important driver of transparency priorities, in which role-based interpretive flexibility is most visible when context activates the professional frame.

\section{Discussion}
\label{sec:discussion}
\subsection{Whose transparency is it?}

A core question this paper asks – whose transparency needs are reflected in the current state of genAI – has a specific empirical answer in our data. It is primarily that of builders, followed by policymakers. While the three stakeholder groups share a common normative foundation for what they need from AI and what they want transparency to achieve, they differ substantially in how well current transparency practices meet their needs. End-users experience mean transparency gaps of 35 points across 13 features, compared to 22 for policymakers and 17 for builders; and only 28\% of end-users report that current practices reflect their goals very well, compared to 53\% of builders and 49\% of policymakers.
The pattern is not random, rather it is consistent with which groups have the most influence over AI systems and transparency artifacts. 

Despite these differences, it is important to note where groups converge. Across needs and transparency goals, the three groups are more similar than different. This convergence is consistent with Suresh et al.’s (2021) argument that interpretability needs should not be tied exclusively to stakeholder roles as higher-level goals can cut across different groups. Our findings provide empirical evidence of this convergence in the context of current genAI transparency, suggesting that the three stakeholder groups do not have fundamentally different conceptions about what transparency means and is for. At the core, transparency matters because it enables people to assess accuracy and calibrate trustworthiness of AI systems and their outputs. This convergence makes the gap pattern that follows more striking, not less. The distributional differences suggest that the extent to which stakeholders' transparency needs are met may depend less on their underlying priorities than on how transparency features are designed, made accessible, and communicated to different groups.

\subsection{The availability gap is a power signal}

The core empirical finding of the paper concerns not where groups disagree on importance, but how they differ in what they perceive as available in current transparency infrastructure. End users have the highest importance-availability gaps, followed by policymakers and then builders. Crucially, this difference is not driven by differences in importance ratings as most features were rated similarly important across groups. It is driven by what they perceive as currently delivered. While there is an overall shortage of transparency across groups, builders perceive more of what they consider important to be available, while end-users perceive much less. This pattern is corroborated by participants’ subjective assessments of how well current practices reflect their goals for what they want transparency to achieve.

Our findings are consistent with critiques of early SCOT's symmetry principle, whereby the pattern of gaps reflects whose priorities have shaped what transparency features get built and for whom. Transparency resources (such as model cards and datasheets), system documentation and benchmarking, and the broader presentations of genAI transparency reflect the technical literacy of builder communities. This pattern is consistent with earlier research which argues that transparency and interpretability methods end up being most understandable to technical audiences versus diverse audiences (Suresh et al, 2021). The one feature approaching stabilization for builders, privacy information (gap=7), is telling. While similarly important for both groups, privacy is perceived as available by 80\% of builders vs 66\% of end-users. Privacy information in current genAI tools is communicated primarily through privacy policies and within settings and information panels. The information and formats are such that technically literate users better know how to find and interpret. Builders’ smaller gap on this feature likely reflects not that privacy tools were built specifically for them, but that existing privacy infrastructure is legible to those who know where to look and how to interpret the information. 

This does not mean these types of transparency approaches and artifacts are “wrong”. Rather, this reflects that they are more easily accessed by builder and technical communities. This is similarly reflected in documentation, whereby the primary tools the field has invested in (e.g., model cards, system cards) are largely unknown to end-users and less useful when encountered, while builders and policymakers are aware of and engage with them at substantially higher rates. These are examples of a broader pattern whereby transparency, even when hypothetically universally available, is predominantly available and legible to those with experience and expertise. This type of transparency does not necessarily create accountability, but rather can serve as a way to maintain positions of power by overwhelming stakeholders with information and documentation that appears transparent while remaining inaccessible or illegible to people (Ananny and Crawford, 2018). 

Importantly, not all gaps reflect distributional differences between groups. Confidence tagging, training data information, and system prompts show major gaps across all three groups. These are less a story about power determining whose needs get met and more about genuinely universal, unmet transparency needs. 

\subsection{General purpose AI and the limits of a single transparency standard}

Current transparency governance largely assumes a convergence model. In this framing, the task is to identify the right standard for transparency, enforce it broadly, and effectively achieve transparency. This is reflected in a policy focus on transparency equating to AI disclosure labels, or industry attention to transparency via documentation. Our findings challenge this assumption in two ways. GenAI, as a general-purpose tool, does not have a single use context and rather offers effectively unbounded use cases, users, and contexts. No single standard can adequately serve the full range of users and contexts. Pursuing one standard will likely result in standards that prioritize the needs of the most centered or powerful group (e.g., builders), while leaving others underserved, creating the perception that transparency is being achieved while being disconnected from meaningful operationalization. 

This does not mean that some aspects of convergence are not possible nor desirable. Our findings show convergence around a shared foundation regarding what transparency should afford. Meanwhile, certain transparency artifacts and practices have emerged that serve groups reasonably well. But these represent a partial and local convergence rather than broad solutions. The story for transparency in genAI is less about arriving at a single right standard, and more about managing permanent plurality, particularly at the practice level. The appropriate governance goal is therefore embracing plurality and cultivating intentional infrastructure that enables meaningful access by different groups. Standards around some aspects, such as documentation requirements or benchmark reporting, can play a useful role, but should be understood as serving particular groups and contexts rather than as universal solutions. 

Critically, end-users also desire more robust governance-related transparency features to understand whether and to what extent to trust an AI system and the organization developing or administering it. In the moment of use, end-users reach for output-level features. However, governance features matter to calibrate one’s trust of a system and the organization behind it. This has implications for researchers. Much of the HCXAI literature has focused on output-level explainability, helping users understand why a model produced a particular response (see for example, \cite{cai2019hello, ghayatrie2026explainable}. Our findings suggest this framing is insufficient. End-users’ needs extend beyond individual outputs to the systems and organizations that produce them. Transparency infrastructure that addresses only what AI systems produce, without addressing who built them, how, and with what accountability, falls short of what people need to calibrate trust meaningfully. This is most clearly illustrated by privacy: all stakeholders expressed protecting privacy and sensitive information as a top need. Privacy concerns are fundamentally about how organizations collect, store, and use personal data – including for what purposes and whose benefit – rather than about any individual AI output. HCXAI and participatory design methods must extend to system-level and governance-related transparency features. 

Finally, in pursuing standards around transparency, language is important. When labeling certain transparency practices, such as AI disclosure, as “transparency legislation”, this risks creating the impression that the transparency challenge has been addressed. This is evident in recent regulation, including Article 50 of the EU AI Act (Regulation (EU) 2024/1689) and California's SB 942 (California AI Transparency Act, 2024). Both focus primarily on ensuring people know when they are interacting with AI or encountering AI-generated content. While important, disclosure of AI involvement or content represents one narrow dimension of what transparency means, and was a priority for a third of policymakers but only 13-15\% of end-users and builders. More precise language (e.g., labeling such legislation as disclosure versus transparency), would better reflect what policies deliver and preserve space for broader transparency regulation and infrastructures. 

\subsection{Transparency as situationally constructed}

The vignette findings add an important dimension, illustrating how transparency priorities are not only role-dependent, but context-dependent. These patterns are consistent with a SCOT-informed understanding of interpretive flexibility as situationally constructed rather than fixed, whereby the same stakeholder group constructs different transparency priorities depending on the context and task structure they are embedded in. In the neutral low-stakes scenario, role-based differences were largely absent: all three groups prioritized explanation, sources, and confidence, and only sources differed by role. Here all participants are functionally end-users, and professional frames stay in the background. When stakes and task structure changed, priorities shifted within the same individuals. End-users’ selection of usage limits rose from 12\% to 44\% in the health scenario, reflecting a desire to calibrate personal trust, and the focus group similarly emphasized limits in high-stakes cases. Builders, asked to choose between systems rather than evaluate an output, moved toward governance features (56\% to 77\% selecting at least one) and away from explanation. Policymakers, whose high-stakes task still involved evaluating an output, shifted from explanation to sources but not toward governance, suggesting that task structure matters alongside stakes. These findings reinforce the argument against uniform transparency standards. If transparency priorities shift substantially with context and task structure even within stakeholder groups, no single standard can adequately serve the full range of situations in which genAI is used.

\subsection{Pathways forward}

Based on these results, we offer three recommendations for researchers, practitioners, and policymakers. First, reorient how transparency is approached. Meaningful transparency in genAI requires shifting from approaching transparency as a technical property to be disclosed or presented, to transparency as a sociotechnical practice to be understood across groups. This includes shifting from a single uniform standard to varied infrastructure that supports different transparency affordances. It also includes shifting from developer-driven artifact production (e.g., static documentation) to processes that redistribute influence over what gets built and how. This does not mean no longer providing documentation, but rather building different types of artifacts that convey similar information in ways that are accessible and digestible to different stakeholders. 

Second, invest in participatory design to develop transparency toolings and infrastructure with underserved groups. Participatory design, particularly with end-users who are most underserved by current transparency practice, is a critical response to the gap patterns we document. These design processes should address transparency not only at the output level, but also at the system and governance level. In particular, our research makes clear that end-users – as well as builders and policymakers – desire clear information about how one’s data is used and protected, as well as broader organizational accountability practices. Privacy policies exist, but are not always accessible to stakeholders and, similar to documentation efforts, require ensuring that approaches are seen as accessible and digestible, which participatory design methods are equipped to support. Part of the opacity in broader accountability mechanisms and governance practices represent opacity in the industry and reflect the lessening transparency being provided about models, their upstream development, and downstream impact \cite{stanford2025foundation}. There are real business interests driving this. In the context of data privacy, opacity can purposefully obscure commercial desires and intention for using personal data for training models, advertising, or other purposes. 

Finally, practitioners can start with the implementation of user-facing transparency features that exhibit consistent gaps and whereby tensions may be minimal. For example, across all stakeholder groups, confidence tags represent a large unmet need. Sources and plain language explanations were also consistently valued by groups and should be provided or continued to be provided. Still, transparency design can remain context-sensitive. Some features, such as usage limits, become more important in high-stakes contexts, suggesting that transparency infrastructure should allow for variation. When implementing transparency features, it is important to recognize that gaps in perceived availability may arise for several reasons: stakeholders may have different levels of access to a feature; relevant information may be available but difficult for certain stakeholders to locate; or stakeholders may encounter the same information but differ in whether they perceive it as satisfying their transparency needs. Understanding these differences can help practitioners determine whether to introduce new features, improve the accessibility and discoverability of existing ones, or communicate information more effectively.

\section{Limitations}
\label{sec:limitations}
There are several limitations. First, gap scores are computed by subtracting perceived availability (a binary self-reported measure) from importance ratings on a Likert scale, combining two methodologically different measures. This approach adapts importance-performance analysis \cite{martilla1977importance}, though unlike standard applications, availability is binary rather than continuous, which is appropriate for measuring perceived access but introduces uncertainty in interpreting gap magnitudes. Second, availability reflects perceived rather than objective availability, which is theoretically appropriate for our research questions. Third and relatedly, the gap analysis is best understood as a stakeholder-centered mapping of perceived transparency needs and gaps, appropriate for a study whose primary aim is to understand how different groups understand and experience transparency versus objectively auditing the transparency tool landscape. Finally, for our focus groups, we were limited to small numbers of respondents that cannot claim to be representative, especially given our practical need to recruit participants who were fluent English-language speakers with reliable Internet access. Finally, our cross-sectional design provides a snapshot of one point in time. 

\section{Conclusion}
\label{sec:conclusion}
What people need from AI and want transparency to achieve is broadly similar across end-users, builders, and policymakers: they want to know genAI systems are accurate and trustworthy, with transparency being a way to help them get there, including through ensuring interactions protect privacy and support reliability. However, current transparency infrastructure serves these groups very differently. Builders experience substantially smaller gaps between transparency needs and what current tools provide, while end-users experience the largest. This pattern is not random. It reflects whose priorities have shaped what transparency artifacts get built and in what form and has consequences for the declining public trust in AI systems. The findings point to three recommendations: (1) reorienting how transparency is approached by recognizing transparency as a sociotechnical practice requiring varied infrastructure, (2) investing in participatory design to develop transparency toolings and infrastructure for end-users and other underserved groups, and (3) implement user-facing transparency features where large, universal gaps persist. Future work can build on this research in several directions. Longitudinal studies can trace how transparency practices evolve as regulation and artifacts evolve. Governance research can inform how regulatory frameworks can accommodate and support interpretive plurality and broader transparency infrastructure. Technical and user design research can ensure transparency features are available and accessible for all, including through participatory design processes with underserved groups.

\section*{Generative AI Usage Statement}
Generative AI (Gemini) was used to assist with grammar and writing fluency.

\section*{Ethical Considerations Statement}
This paper examines how end-users, AI builders, and policymakers understand and experience transparency in generative AI. We hope these findings support more meaningful and equitable transparency practices, but we acknowledge they could be misread in several ways. First, the finding that stakeholders share a common understanding of what transparency should achieve could be taken as support for a single, uniform transparency standard. We caution against this reading. Our results show that while goals converge, needs and experiences differ substantially by role and context, and no single standard is likely to serve the full range of genAI users and uses. Second, the finding that end-users are largely unaware of model and system cards, and find them less useful, could be used to argue that such documentation is unnecessary or to justify reduced disclosure. Our results do not suggest that documentation should be scaled back. Rather, the information it contains should also be conveyed in forms that are accessible and legible to people beyond technical communities. Finally, we recognize that participatory design with underserved groups can become extractive or tokenistic. We argue that participatory processes should meaningfully redistribute influence over what transparency infrastructure gets built, rather than serving as a consultative step.


% ...

\bibliography{acmart}

\begin{appendices}

  \input{appendix}

\end{appendices}

\end{document}

%% file: appendix.tex

\clearpage
\raggedbottom
\appendix
\section*{Appendix}

\setcounter{figure}{0}
\setcounter{table}{0}
\renewcommand{\thefigure}{A\arabic{figure}}
\renewcommand{\thetable}{A\arabic{table}}


\subsection*{Supplementary Figures}
\label{app:4}

\begin{figure}[H]
  \centering
  \includegraphics[width=0.5\linewidth]{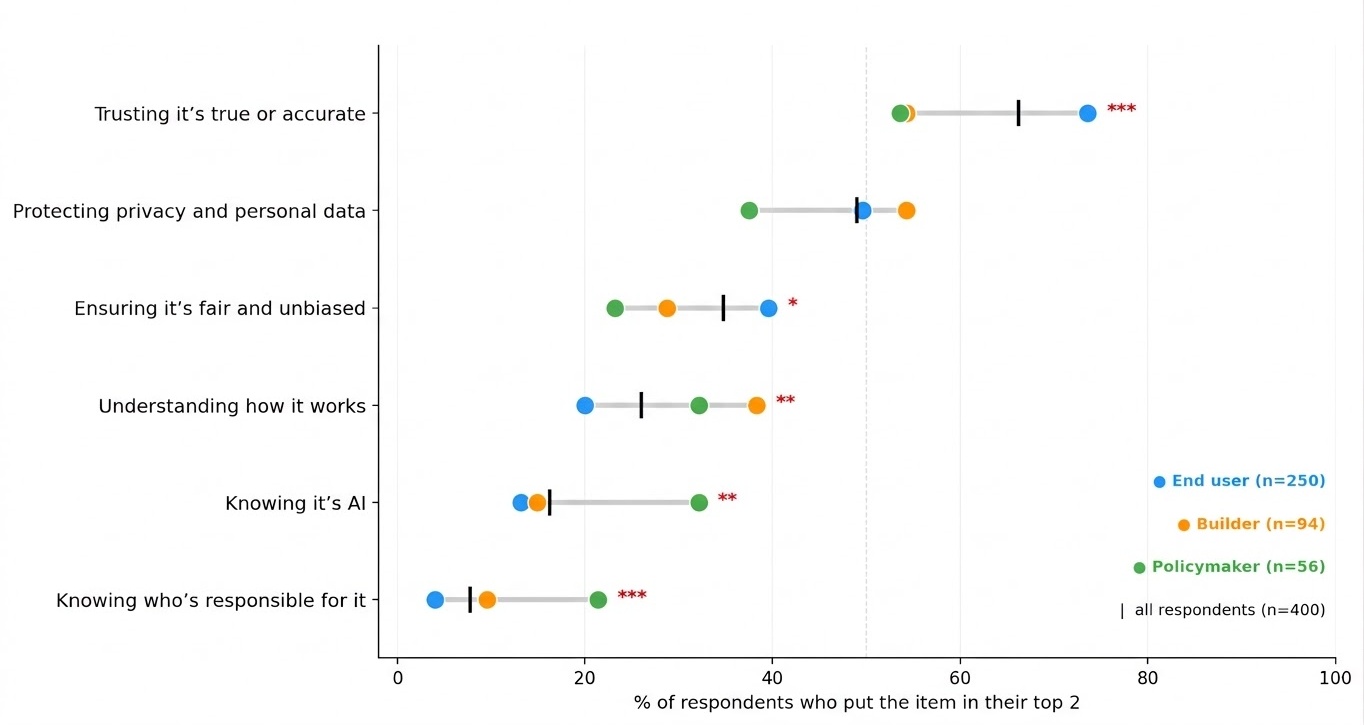}
  \caption{What matters most, by stakeholder role.}
  \label{fig:app-matters}
  \vspace{2pt}
  {\scriptsize\raggedright \textit{Note.} Respondents selected exactly 2 of 6 items in response to ``When you use generative AI tools, which of the following aspects matter most to you?'' Stars mark an omnibus chi-square test of independence across the three roles for each item (role $\times$ selected/not selected): * $p < .05$, ** $p < .01$, *** $p < .001$. Significant omnibus differences were followed up with pairwise Fisher's exact tests reported here (accurate: EU vs.\ B $p < .001$, EU vs.\ PM $p < .001$; fairness: EU vs.\ PM $p < .05$; how it works: B vs.\ EU $p < .01$; knowing it's AI: PM vs.\ EU $p < .01$, PM vs.\ B $p < .05$; responsibility: PM vs.\ EU $p < .001$). EU = end-user, B = builder, PM = policymaker.\par}
\end{figure}

\begin{figure}[H]
  \centering
  \includegraphics[width=0.5\linewidth]{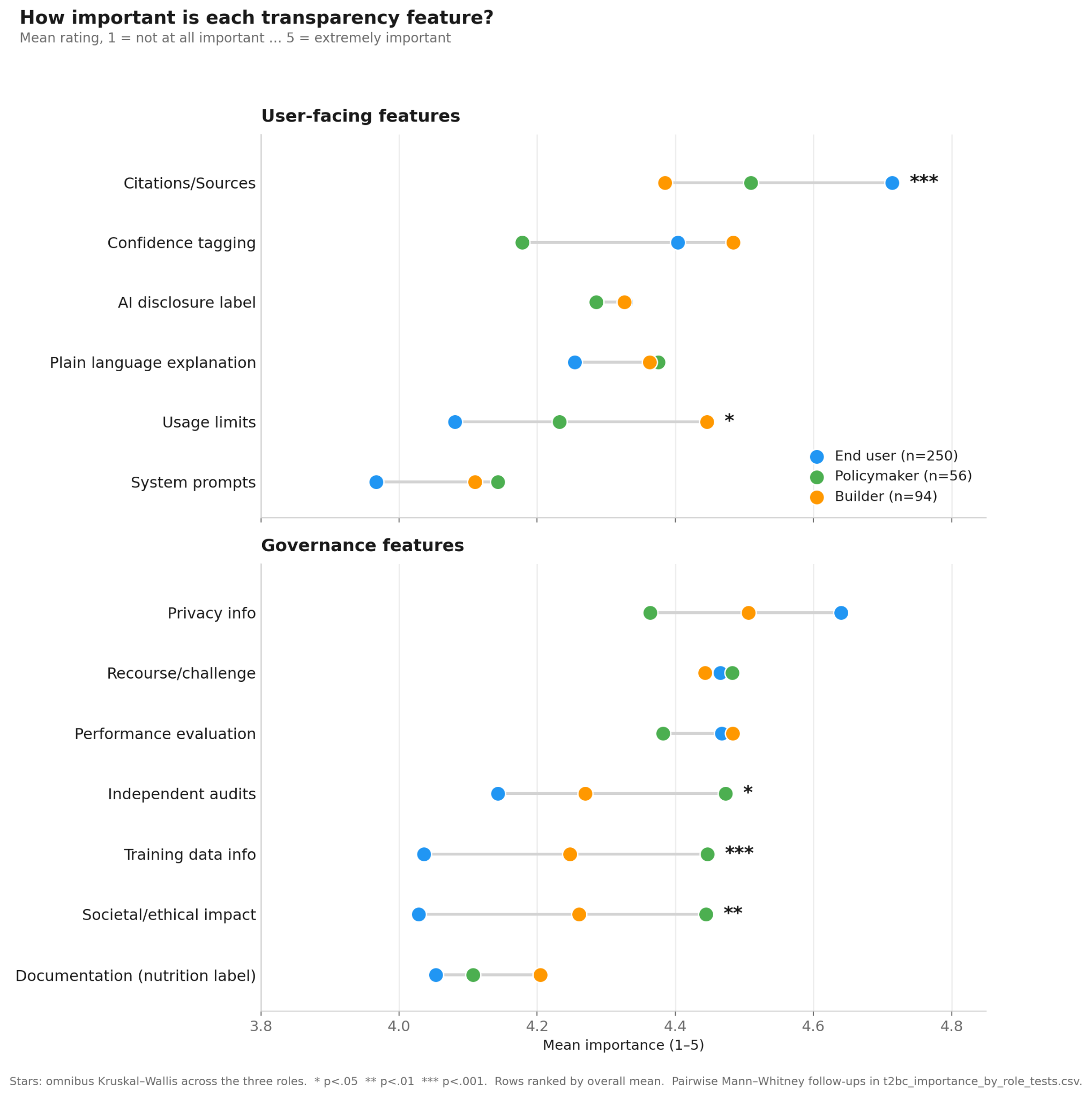}
  \caption{Importance of transparency features, by stakeholder group.}
  \label{fig:app-importance}
  \vspace{2pt}
  {\scriptsize\raggedright \textit{Note.} This shows the mean importance rating (1 = not at all important, 5 = very important) for each transparency feature by stakeholder. Stars on the row labels report an omnibus Kruskal--Wallis test of whether the three role distributions differ (* $p < .05$, ** $p < .01$, *** $p < .001$), applied per feature with no correction for multiple comparisons across the 13 features. Where the omnibus test was significant, pairwise Mann--Whitney U tests were run as follow-ups. For AI disclosure label, the end-user mean (4.33) is identical to the builder mean, so the two markers coincide with the builder data point visible.\par}
\end{figure}

\begin{figure}[H]
  \centering
  \includegraphics[width=0.75\linewidth]{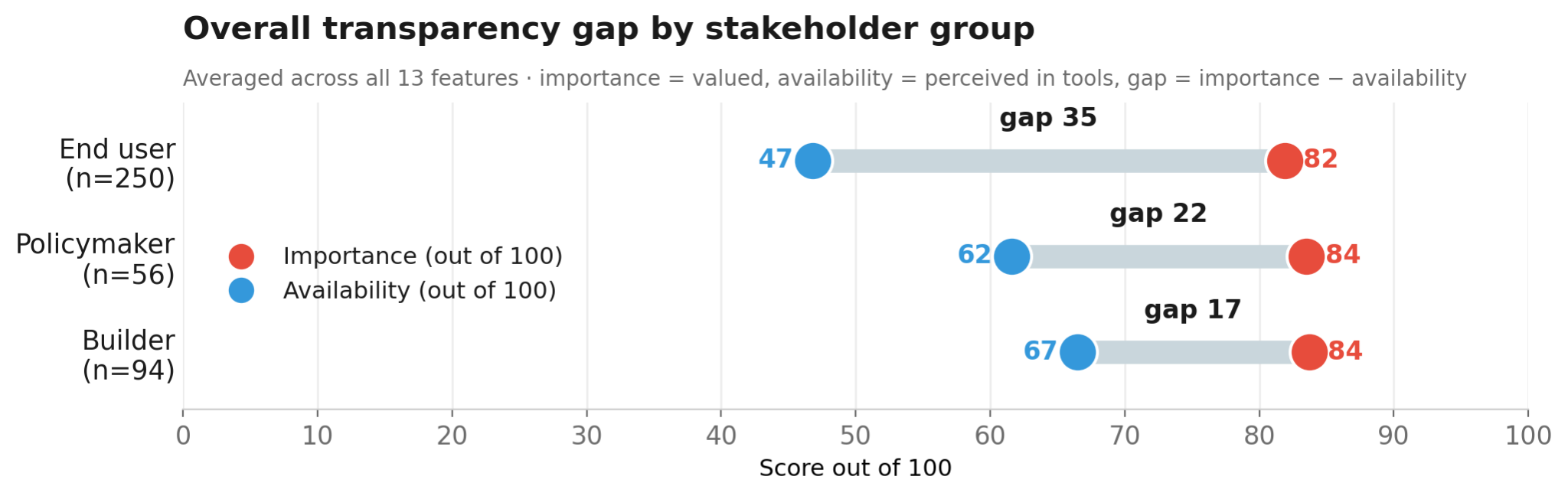}
  \caption{Overall transparency gaps, by stakeholder group.}
  \label{fig:app-gaps}
  \vspace{2pt}
  {\scriptsize\raggedright \textit{Note.} Scores are averaged across all 13 transparency features. Importance is the mean 1--5 rating rescaled to 0--100; availability is the percentage reporting that their latest or most-used generative AI tool provided the feature; gap = importance $-$ availability.\par}
\end{figure}

\begin{figure}[H]
  \centering
  \includegraphics[width=0.5\linewidth]{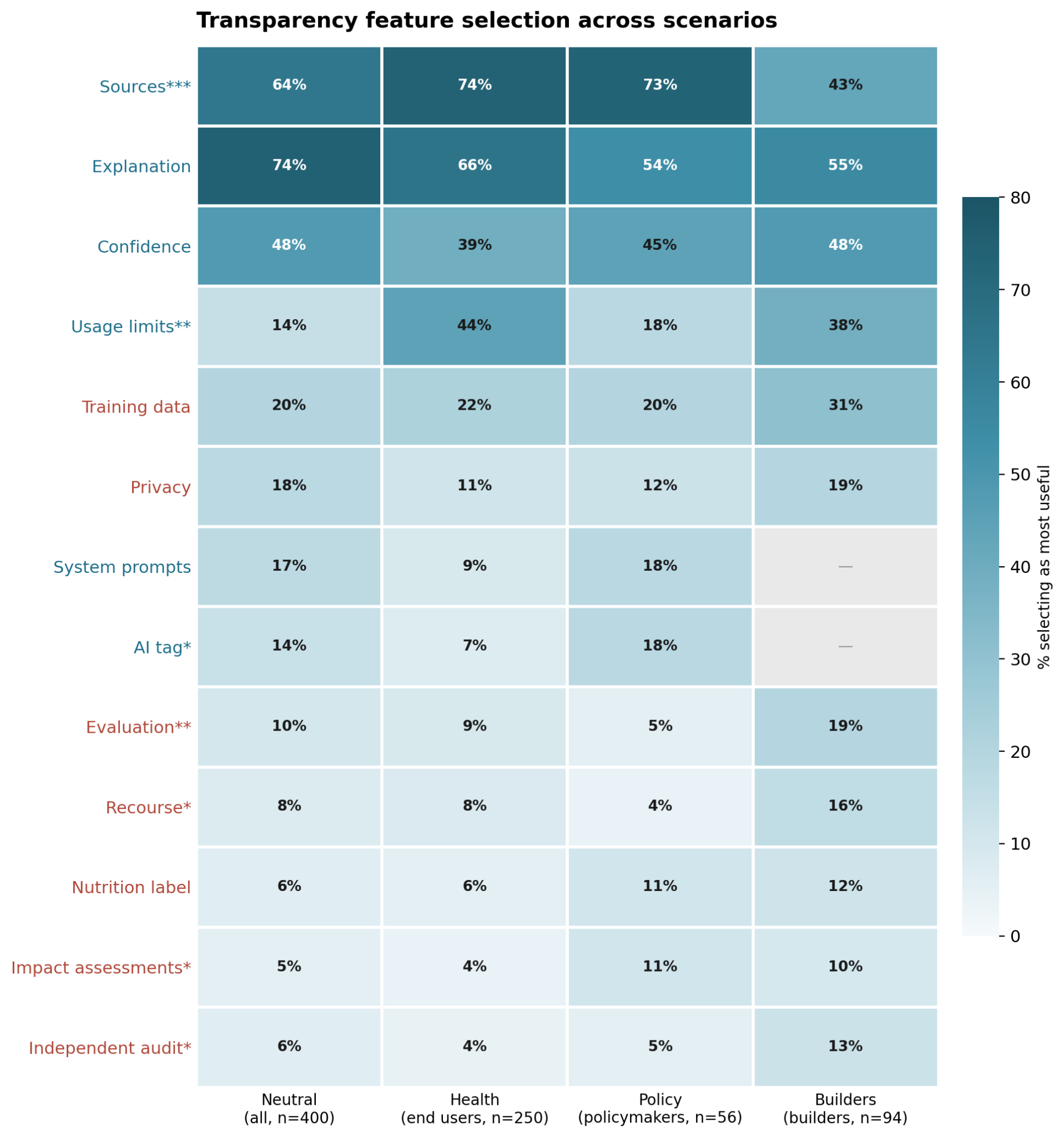}
  \caption{Percent of respondents selecting transparency features, across scenarios.}
  \label{fig:app-scenarios}
  \vspace{2pt}
  {\scriptsize\raggedright \textit{Note.} Percentage choosing each feature among their three most useful. Blue = user-facing, red = governance. Contextual scenarios were shown only to their own stakeholder group. Because differences between columns combine context and role, we rely primarily on within-person comparisons, testing each group's shift from the neutral to its role-specific scenario with exact McNemar tests. Stars report chi-square tests across the three contextual scenarios (* $p < .05$, ** $p < .01$, *** $p < .001$). Dashes = not offered in scenario option.\par}
\end{figure}

\subsection*{Appendix 1a. Survey Recruitment and Inclusion Criteria Details}
\label{app:1a}

The goal of this preregistered survey was to recruit participants representing three stakeholder groups in AI development and governance: end-users, builders, and policymakers. Eligibility was determined by role. End users were required to use generative AI tools in their primary role; builders were required to have experience in AI product management or technical AI development; and policymakers were required to have experience informing or advising on AI regulation or governance. All participants were required to be adults (age of majority in their country of residence), speak English, and be located in one of the countries specified in the IRB-approved recruitment protocol spanning North America, South America, Europe, Africa, Asia, and Oceania. Participants from Singapore were required to be at least 21 years old and participants from New Zealand at least 20 years old, consistent with the age of majority in those countries.

Participants were recruited through a combination of Prolific and targeted outreach through the research team's professional networks. Recruitment strategies varied by stakeholder group. End-users were recruited primarily through Prolific, with the goal of obtaining geographic diversity across countries and regions specified in the IRB-approved recruitment protocol. Builders and policymakers were recruited primarily through targeted outreach to individuals within the research team's professional networks, supplemented by recruitment through Prolific. This approach was intended to reach participants with relevant experience in AI development, deployment, and governance. The resulting sample comprised 400 respondents: 250 end-users, 94 builders, 56 policymakers, and 4 respondents classified as other. Of these, 290 were recruited through Prolific and 110 through professional-network-based outreach.

\subsection*{Appendix 1b. Survey Respondent Demographics}
\label{app:1b}

\begin{table}[H]
  \centering
  \caption{Demographic characteristics of survey respondents ($N = 400$).}
  \label{tab:demographics}
  \footnotesize  
  \renewcommand{\arraystretch}{0.4}
  \begin{tabular}{@{}lrr@{}}
    \toprule
    \textbf{Characteristic} & \textbf{n} & \textbf{\%} \\
    \midrule
    \multicolumn{3}{@{}l}{\textbf{Gender}} \\
    \quad Male       & 210 & 53\% \\
    \quad Female     & 185 & 46\% \\
    \quad Non-binary &   5 &  1\% \\
    \addlinespace
    \multicolumn{3}{@{}l}{\textbf{Age}} \\
    \quad 18--24 &  72 & 18\% \\
    \quad 25--34 & 187 & 47\% \\
    \quad 35--44 &  88 & 22\% \\
    \quad 45--54 &  37 &  9\% \\
    \quad 55--64 &  10 &  2\% \\
    \quad 65+    &   6 &  2\% \\
    \addlinespace
    \multicolumn{3}{@{}l}{\textbf{Race / ethnicity}$^{1}$} \\
    \quad White                    & 169 & 42\% \\
    \quad Black / African American & 156 & 39\% \\
    \quad Latino / Hispanic        &  33 &  8\% \\
    \quad East Asian               &  25 &  6\% \\
    \quad South Asian              &  20 &  5\% \\
    \quad Other                    &   9 &  2\% \\
    \quad Indigenous               &   6 &  2\% \\
    \quad Pacific Islander         &   1 &  0\% \\
    \addlinespace
    \multicolumn{3}{@{}l}{\textbf{Country income group}} \\
    \quad High-income            & 241 & 60\% \\
    \quad Upper-middle           & 147 & 37\% \\
    \quad Low- and middle-income &  12 &  3\% \\
    \addlinespace
    \multicolumn{3}{@{}l}{\textbf{Country of residence}$^{2}$} \\
    \quad United States                        & 144 & 36\% \\
    \quad South Africa                         & 126 & 32\% \\
    \quad Canada                               &  21 &  5\% \\
    \quad United Kingdom                       &  19 &  5\% \\
    \quad Mexico                               &  15 &  4\% \\
    \quad India                                &   9 &  2\% \\
    \quad Poland                               &   9 &  2\% \\
    \quad Australia                            &   8 &  2\% \\
    \quad United States Minor Outlying Islands &   7 &  2\% \\
    \quad Chile                                &   6 &  2\% \\
    \quad Other (17 countries)                 &  36 &  9\% \\
    \addlinespace
    \multicolumn{3}{@{}l}{\textbf{AI / coding experience}} \\
    \quad No coding experience      & 114 & 28\% \\
    \quad Reads code, rarely writes & 105 & 26\% \\
    \quad Applies AI methods        &  57 & 14\% \\
    \quad Writes/debugs confidently &  55 & 14\% \\
    \quad Expert (builds models)    &  41 & 10\% \\
    \quad Adapts/trains models      &  28 &  7\% \\
    \addlinespace
    \multicolumn{3}{@{}l}{\textbf{Gen-AI usage frequency}} \\
    \quad Daily (many times)  & 137 & 34\% \\
    \quad Weekly              & 121 & 30\% \\
    \quad Daily (once/twice)  &  99 & 25\% \\
    \quad Monthly             &  35 &  9\% \\
    \quad Rarely              &   8 &  2\% \\
    \bottomrule
  \end{tabular}

  \vspace{4pt}
  \begin{minipage}{0.75\linewidth}
    \footnotesize
    $^{1}$Race/ethnicity was a select-all-that-apply item; percentages therefore sum to more than 100\%.\\
    $^{2}$Respondents resided in 27 countries in total. The ten most represented countries are shown individually; the remaining 17 countries are grouped as ``Other.''
  \end{minipage}
\end{table}

\subsection*{Appendix 2. Structured Focus Group Activity Descriptions and Illustrative Outputs}
\label{app:2}

\paragraph{Activity 1. Transparency Needs and What's Important}

Participants were divided into small groups, each with their own shared digital board in Canva (an online collaborative whiteboard). Each board contained a set of virtual sticky notes. Participants individually added sticky notes in response to the brainstorm prompt, then worked together as a group to organize and vote on them.

\textit{Individual brainstorm prompt (5 minutes):} ``When you think about transparency in generative AI systems, what specific things do you want or need to know? And why?''

\textit{Small group discussion (15 minutes):} Participants organized their sticky notes into categories, created labels, and discussed together. Groups then briefly discussed what values drive these needs using the following value word bank: Fairness/Equity, Accountability, Privacy, Autonomy/User Control, Trustworthiness, Safety/Reliability, Accessibility/Inclusion.

\textit{Full group discussion:} One person from each group shared out in the large group and discussed themes all together.

\begin{figure}[htbp]
  \centering
  \includegraphics[width=0.75\linewidth]{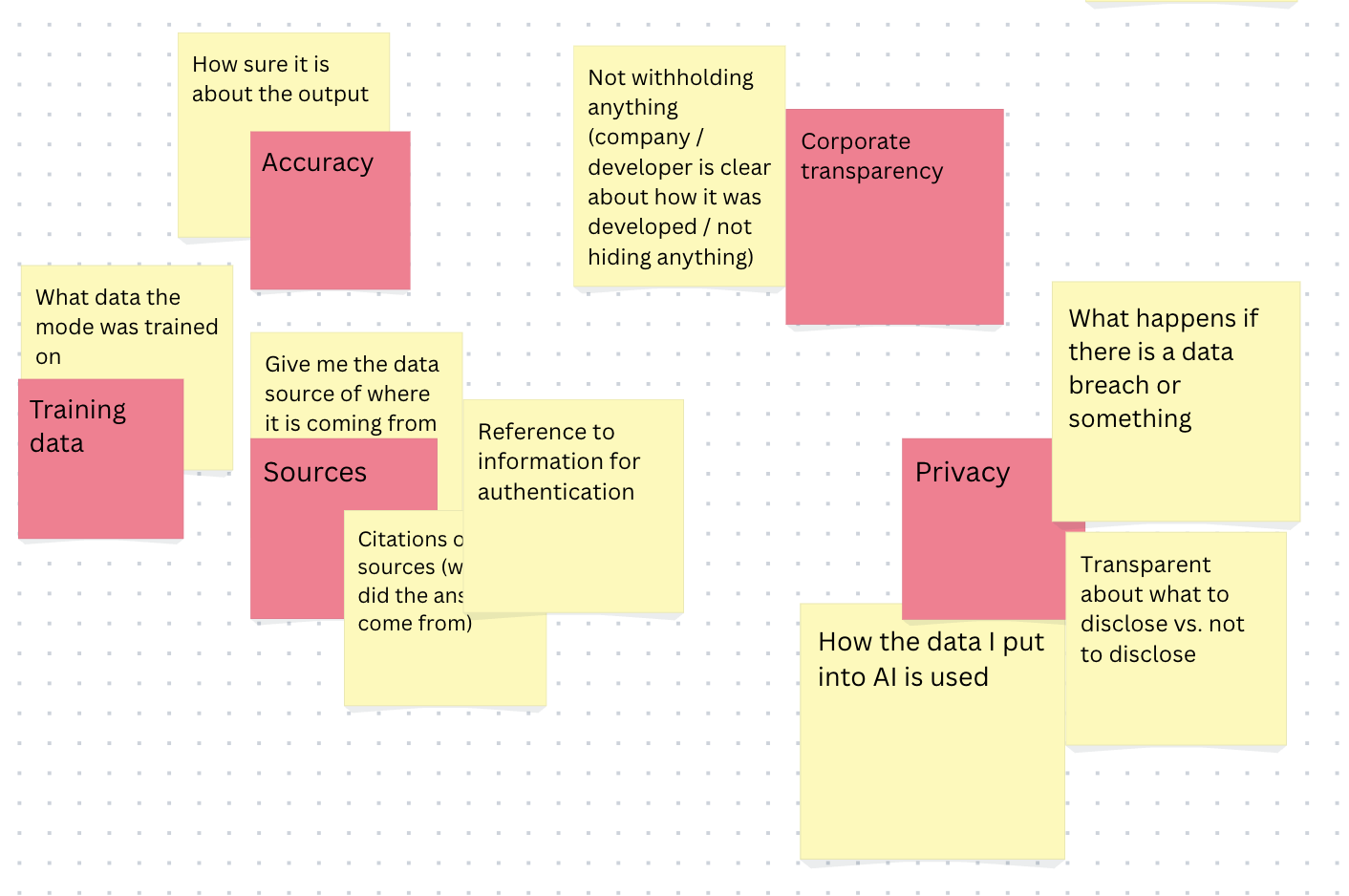}
  \caption{Activity 1. Illustrative example of one group's transparency needs and what's important.}
  \label{fig:app-activity1}
\end{figure}

\paragraph{Activity 2. Contextual Transparency Needs}

\begin{figure}[htbp]
  \centering
  \includegraphics[width=\linewidth]{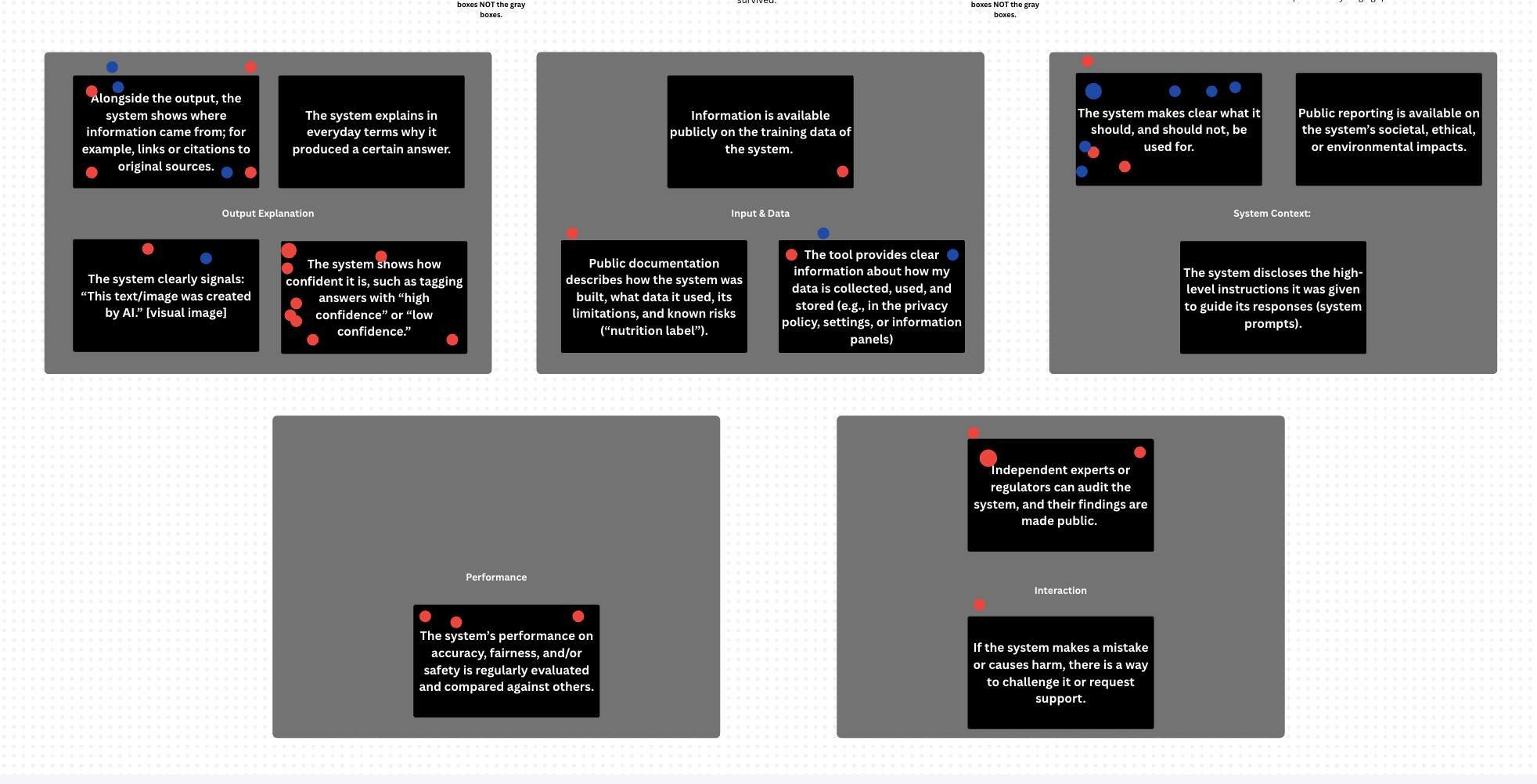}
  \caption{Activity 2. Transparency features to vote on, along with one group's voting patterns.}
  \label{fig:app-activity2}
\end{figure}

Participants were divided into small groups, each with their own shared digital board in Canva. Each board displayed a vignette and the 13 transparency features in boxes (see Figure~\ref{fig:app-activity2}). Participants read vignettes depicting generative AI interactions in different contexts and, for each vignette, completed two rounds of individual dot voting on their group's Canva board, followed by a small group discussion. Participants completed as many vignettes (up to 3) as they were able to in the time allotted.

\textit{Round 1 -- Most Important (1 minute):} ``If you could improve AI systems in only a few of the transparency features below, which would most help you accomplish your goals?'' Each participant placed up to 7 red dots on the transparency features they considered most important. Multiple dots could be placed on a single feature.

\textit{Round 2 -- Currently Provided (1 minute):} ``When you use AI systems today, which of these transparency features do you actually see provided in practice?'' Each participant placed up to 7 blue dots on the transparency features where they most commonly see relevant information provided in practice. Multiple dots could be placed on a single feature.

\textit{Small Group Discussion (after all vignettes):} ``What surprises you about the gap between red (desires/needs) and blue (what's provided)? Are there areas where there are particularly large gaps?''

\medskip
\noindent\fbox{\parbox{\dimexpr\linewidth-2\fboxsep-2\fboxrule\relax}{%
\textbf{Vignette 1: Healthcare (High Stakes)}\par\smallskip
Maria is using an AI chatbot to ask about a pain in her chest she has had for the last fifteen minutes. Unsure if it's serious, she explains her symptoms and asks the chatbot: ``What would you recommend I do based on my symptoms?'' The chatbot responds, ``Your symptoms do not appear to be an emergency. You may be experiencing stress or indigestion. Try resting and monitoring the pain. Seek medical care if symptoms persist.'' Maria is uncertain, because she cannot tell how the chatbot reached this conclusion or how urgent her situation might be.\par\smallskip
Maria's son, who works as a nurse, comes home and Maria tells him what happened. Her son is alarmed by Maria's symptoms and insists on driving her to the emergency room. On the way there, Maria's symptoms worsen and by the time she arrives it is clear she is having a heart attack. The doctors manage to give her treatment in time. One doctor says to her, ``You're lucky you came when you did. A few more minutes and you might not have survived.''}}

\medskip
\noindent\fbox{\parbox{\dimexpr\linewidth-2\fboxsep-2\fboxrule\relax}{%
\textbf{Vignette 2: Shopping (Low Stakes)}\par\smallskip
Jordan is using an AI chatbot on a shopping website to find running shoes he can buy online. He types, ``I'm looking for decent running shoes, but I want to stay under \$100.'' The chatbot responds, ``I recommend this premium pair for \$140, it's an excellent match for you.'' The shoes look appealing, and the chatbot's recommendation seems confident, but Jordan hesitates. He wonders why this option was chosen, whether the recommendation is based on his preferences or the company's interests, and whether advertising or sponsorship might be influencing the result.}}

\medskip
\noindent\fbox{\parbox{\dimexpr\linewidth-2\fboxsep-2\fboxrule\relax}{%
\textbf{Vignette 3: Social Media \& AI Images (High Stakes)}\par\smallskip
Scrolling through social media, Aisha sees a striking image of a snowbank that is higher than a small house in a nearby city. She heard about record snowfall on the evening news but the amount of snow is more than anything she has seen or heard yet.\par\smallskip
Her elderly parents live in that city and she begins to worry for their safety. She gives them a phone call but no one answers. Concerned for their safety, she decides she will drive to see them tomorrow if they do not call her back. She wakes up the next day and prepares for the drive. Just as she is about to leave, her parents call her back. They report that everything is fine and that the snowfall is much less than in the photo. Aisha is relieved to hear everyone is safe but is confused about the photos. After a quick online search, Aisha finds that the images she saw were generated by AI, but nothing in the post had made that clear.}}
\medskip

\subsection*{Appendix 3. Survey Vignettes (Scenarios)}
\label{app:3}

\textit{About:} All respondents receive one neutral vignette, then each type of participant receives another that is specific to their role (e.g., policymaker, builder).

\textit{Prompt:} ``You will now receive two scenarios that reflect a real-world use of a generative AI tool and requests for more transparency.''

\paragraph{Vignette 1. Low-stakes {[All respondents]}}

You are using a generative AI chatbot to get reliable information on a factual question that matters for an upcoming decision. You ask: ``Where is the Magnetic North Pole located?''

The chatbot responds: ``The location of the Magnetic North Pole changes significantly each year, currently moving towards Siberia.''

You asked for additional transparency/explanation about this answer and received the following options. Please rank them based on which response is most valuable to you (1 = least useful, 10 = most useful).

\begin{itemize}
  \item AI tag -- ``This text was generated by AI.''
  \item Explanation -- ``I obtained this answer by reviewing multiple sources and selecting the one with the strongest scientific consensus. My data goes up to 2024, so it may be outdated.''
  \item Confidence -- ``High confidence: I am 96\% confident that this answer is correct.''
  \item Limits -- ``This system should only be used for educational study support, not for making final academic claims.''
  \item Sources -- ``I obtained this answer from the following source: World Data Center for Geomagnetism, Kyoto (2024). [link]''
  \item System prompts -- ``I have been instructed to provide factual answers unless my confidence is below 90\%. You can see the full set of instructions I was given here: [link].''
  \item Training data -- ``This system was trained on large collections of scientific texts and peer-reviewed journals. You can read more about the datasets here: [link].''
  \item Documentation (``nutrition label'') -- ``There is a public document describing how the system was built, what data it used, its limitations, and risks: [link].''
  \item Impact assessments -- ``Public reports are available on the system's societal, ethical, and environmental impacts: [link].''
  \item Evaluation -- ``This system's accuracy and fairness are regularly compared against other AI systems. Reports are available here: [link].''
  \item Audit -- ``Independent experts test this system for accuracy, bias, and safety. Their findings are published here: [link].''
  \item Privacy -- The tool provides clear information about how your data is collected, used, and stored (e.g., in the privacy policy, settings, or information panels).
  \item Recourse -- ``If you believe this answer is incorrect or harmful, you can submit feedback or request a review here: [link].''
\end{itemize}

\textit{Explain why the top ranked response(s) are most valuable to you?}

\paragraph{Vignette 2. High-stakes {[End users only]}}

You have been experiencing headaches for the last four hours and want to better understand some potential causes for the headache before you seek treatment. You decide to ask a generative AI chat bot about your symptoms. You ask the following question: ``What are some potential causes of a headache lasting more than four hours?''

You receive the following response: ``There are many potential causes. Your headache might either be caused by a primary headache disorder (migraines, tension headaches, cluster headaches) or a secondary cause (dehydration, sleep issues, caffeine withdrawal, sinus congestion, eye strain). These are just some of the more common causes but the list is not exhaustive.''

You asked for additional transparency/explanation about the answer and received several options. Please rank the outputs based on which response is most valuable to you (1 is least useful, 10 is most useful).

\begin{itemize}
  \item AI tag -- ``This text was generated by AI.''
  \item Sources -- ``I obtained this answer from the following source, which is widely cited in the medical literature: Robbins, M. (2021). \textit{Diagnosis and management of headache.} JAMA.''
  \item Explanation -- ``I reviewed medical articles and health websites to list the most common causes. There is a chance this is incomplete or incorrect, since many other causes exist.''
  \item Confidence -- ``High confidence: I am 96\% confident that this answer is correct.''
  \item Limits -- ``This system should be used only for general information and not for diagnosis or treatment. You should consult a medical professional for care.''
  \item System prompts -- ``I was instructed to provide factual answers unless my confidence is below 90\%. You can read my instructions here: [link].''
  \item Training data -- ``This system was trained on large collections of peer-reviewed medical texts and health resources. More details are available here: [link].''
  \item Documentation (``nutrition label'') -- ``A public document explains how this system was built, what data it used, its limitations, and known risks: [link].''
  \item Impact assessments -- ``Reports are available on potential societal or ethical impacts of using this system (e.g., risks of bias in diagnosis suggestions): [link].''
  \item Evaluation -- ``The system's accuracy and safety are regularly compared with other medical AI tools. Reports are available here: [link].''
  \item Audit -- ``Independent experts test this system for accuracy, bias, and safety. Results are published here: [link].''
  \item Privacy -- The tool provides clear information about how my data is collected, used, and stored (e.g., in the privacy policy, settings, or information panels).
  \item Recourse -- ``If you believe this answer is incorrect or harmful, you can report the issue or request review here: [link].''
\end{itemize}

\textit{Explain why the top ranked response(s) are most valuable to you?}

\paragraph{Vignette 3. High-stakes {[Builders only]}}

You are on a product team that is considering using a generative AI model to power a new chatbot feature. The chatbot will interact directly with users, answering their questions and providing support. Your team is comparing different models. Some models offer more transparency opportunities in communicating to users and some offer more transparency about how they are built and governed. You want to choose the product that will provide your customers with the best and most trustworthy answer.

Please rank what kinds of transparency are most valuable to you when considering what model to use (1 is least useful, 10 is most useful).

\begin{itemize}
  \item Explanation -- The model supports features your team can use to provide everyday explanations of why it produced a certain answer.
  \item Confidence -- The model can express how confident it is in its answers, which your team could surface to users.
  \item Limits -- The model comes with clear statements about what it should, and should not, be used for---helping your team set safe boundaries.
  \item Sources -- The model can provide citations or references for its outputs, which your team could integrate into the product.
  \item Training data -- Information is available publicly on what kinds of data the model was trained on.
  \item Documentation (``nutrition label'') -- Public documentation exists describing how the system was built, what data it used, its limitations, and known risks.
  \item Evaluation -- Reports compare this model's accuracy, fairness, and safety against other models.
  \item Audit -- Independent experts or regulators test the system for accuracy, bias, or safety, and publish the results.
  \item Impact assessments -- Public reporting exists on the system's societal, ethical, or environmental impacts.
  \item Privacy -- The tool provides clear information about how my data is collected, used, and stored (e.g., in the privacy policy, settings, or information panels).
  \item Recourse -- If the model makes serious errors or causes problems in your app, there is a process to escalate, challenge, or request support from the provider.
  \item None of these are valuable to me.
\end{itemize}

\textit{Explain why the top ranked response(s) are most valuable to you?}

\paragraph{Vignette 3. High-stakes {[Policymakers only]}}

You are a policymaker preparing a briefing on whether new transparency rules for generative AI should be introduced in your region. As part of your research, you ask a generative AI system: ``What are the current standards on transparency in AI?''

The system responds: ``Several organizations have developed transparency frameworks for AI. For example, the EU's AI Act requires documentation of training data, risk assessments, and system performance disclosures, while the U.S.\ NIST AI Risk Management Framework highlights transparency as a core principle for trustworthy AI. Most standards emphasize clear communication with users, documentation of system design, and accountability mechanisms.''

You asked for additional transparency/explanation about the answer and received several options. Please rank the outputs based on which response is most valuable to you (1 is least useful, 10 is most useful):

\begin{itemize}
  \item AI tag -- ``This text was generated by AI.''
  \item Sources -- ``I obtained this answer from the following sources, which I chose because they are widely referenced in AI governance: European Union AI Act (2024); NIST AI Risk Management Framework (2023).''
  \item Explanation -- ``I reviewed policy documents and academic literature on AI regulation to identify the most widely cited standards. There is a chance this is incomplete, since new guidelines are emerging rapidly.''
  \item Confidence -- ``High confidence: I am 92\% confident this summary reflects the most current leading standards.''
  \item Limits -- ``This system should be used for general background and not as a substitute for legal advice or official regulatory guidance.''
  \item System prompts -- ``I was instructed to provide factual summaries of AI standards, avoid speculation, and indicate uncertainty when applicable. You can read my instructions here: [link].''
  \item Training data -- ``This system was trained on a mixture of public policy documents, academic articles, and news sources. More details are available here: [link].''
  \item Documentation (``nutrition label'') -- ``A public document explains how this system was built, what data it used, its limitations, and known risks: [link].''
  \item Impact assessments -- ``Reports are available on potential societal, ethical, and governance impacts of using this system in policymaking: [link].''
  \item Evaluation -- ``This system's performance on accuracy and fairness is regularly compared with other AI models. Reports are available here: [link].''
  \item Audit -- ``Independent experts and regulators test this system for bias, reliability, and safety. Results are published here: [link].''
  \item Privacy -- The tool provides clear information about how your data is collected, used, and stored (e.g., in the privacy policy, settings, or information panels).
  \item Recourse -- ``If you believe this answer is incorrect or harmful, you can report the issue or request review here: [link].''
  \item None of these are useful to me -- I prefer just the answer.
\end{itemize}

\textit{Explain why the top ranked response(s) are most valuable to you?}